# AI driven Shadow Model Detection in AgroPV Farms

**Sai Paavan Kumar Dornadula**
School of Electronics Engineering
Vellore Institute of Technology, Chennai
Chennai,India
dornadulasai.paavan2019@vitstudent.ac.in

**Pascal Brunet**
Naldeo Technologies
Industries
Tarnos, Nouvelle-Aquitaine, France
pascal.brunet@naldeo.com

**Dr. Susan Elias**
School of Electronics Engineering
Vellore Institute of Technology, Chennai
Chennai,India
susan.elias@vit.ac.in

## Abstract

Agro-photovoltaic (APV) is a burgeoning farming practice that combines agricultural activities and solar photovoltaic projects (PV) within the same area. This market is projected to experience significant growth in the next few years, with investments of approximately 9 billion US dollars expected in 2030, accounting for over 5% of the total photovoltaic investment. In certain countries, APV may even become the primary market for photovoltaic.

The identification of shadows is a crucial aspect of APV research because shadows significantly impact several phenomena in the APV environment, including plant growth, microclimate, and evapotranspiration. Shadow detection can be highly beneficial, enabling a better understanding of the environment and ensuring a fair distribution of solar energy between PV and agricultural activities. In this study, we focus on using state-of-the-art CNN and GAN-based neural networks to identify shadows in agro-PV farms. CNN-based techniques can categorize pixels as either shadow or non-shadow by extracting features from input images using convolutional layers, while GAN-based networks can produce synthetic shadow images and differentiate between fabricated and genuine images. Our case studies demonstrate the effectiveness of these techniques in detecting shadows in agro-PV farms, but we also acknowledge the challenges that need to be addressed, such as the impact of partial shadowing from moving objects like animals or people, as well as the need for real-time monitoring. We suggest that future research should focus on developing more sophisticated neural network-based shadow detection algorithms and integrating shadow detection with other control systems for agro-PV farms. Overall, this study emphasizes how important shadow detection is for increasing the productivity and profitability of agro-PV farms while offering environmental and social benefits that support the plants, soil, and farmers.

## 1 Introduction

An intriguing and relatively new method of producing food and energy sustainably is agro-photovoltaic (PV) farming. Using solar panels to generate electricity while providing a shaded environment for crops to flourish, it blends the ideas of agriculture and photovoltaic power generation[1, 2]. However, the crops and other objects' shadows have an impact on how well the solar panels perform, which lowers energy output. As a result, shadow detection is crucial to ensuring the best possible performance of PV systems in agro-PV farms[3,4,5,6]. The major algorithms used in shadow detection approaches for PV systems are those for image processing and machine learning. Using image processing techniques, images are divided into shadow and non-shadow regions, and then characteristics are extracted to aid discriminate between the two. On the other hand, machine learning algorithms employ data-driven strategies to learn from a collection of photographs and anticipate the presence of shadows. Convolutional Neural Networks (CNNs) have become a prominent option for PV system shadow detection. Deep learning systems known as CNNs are capable of learning to recognize intricate patterns in photos. They have been utilized in numerous investigations and have demonstrated remarkable accuracy in the detection of shadows. The detection of shadows in agro-PV fields has also been investigated using Generative Adversarial Networks (GANs). The objective is to train a discriminator to classify between authentic and synthetic images, GANs are employed to create synthetic images that are comparable to real ones. These networks exhibit promising results in PV systems being able to detect shadows. Although shadow detection methods have advanced, there are still issues that need to be resolved. The effect of partial shade by moving objects, such as animals or workers, on the energy production of PV systems is one of the difficulties. This is especially important in agro-PV farms because of the constant movement of animals and people. Real-time shadow monitoring and solar panel placement adjustments present another difficulty. Continuous environmental monitoring and dynamic PV system control are necessary for this. With a focus on agro-PV farms, we intend to give a thorough analysis of the most recent methods for detecting shadows on PV systems in this research study. We'll discuss the methods that already exist that use machine learning and image processing. Additionally, we will contrast and assess the effectiveness of various machine learning methods, such as CNNs and GANs, for shadow detection in agro-PV fields. We will also draw attention to the difficulties and openings for further study in this area. The primary goal of this study is to present a summary of the SOA in PV system shadow detection techniques, with an emphasis on agro-PV farms. Understanding the benefits and drawbacks of the various strategies will enable researchers and practitioners in the field of agro-photovoltaic farming to choose which ones to employ. Additionally, this review will aid in the creation of more precise and effective shadow detecting methods, enhancing the dependability and performance of agro-PV farms. Overall, this study has the potential to significantly improve the sustainability of food and energy production, which will ultimately improve the health of our planet and its inhabitants.

## 2  Related Work

The problem of identifying shadows in a single image has been extensively studied.. Various techniques have been developed for shadow detection, including traditional and deep learning-based approaches. Recent studies have explored methods such as geometry-based techniques [7][8], chromaticity-dependent techniques [9][10][11][12], physical methods [13][14][15], edge-based methods [16], and texture-based methods [17][18][19]. However, these conventional techniques tend to beslower and less accurate in detecting shadows.. Data-driven methods based on building classifiers from small, annotated datasets [20] have recently demonstrated greater success. Vicente et al., An example of using a leave-one-out-based multi-kernel LS-SVM includes optimizing the LS-SVM through multiple kernel functions and assessing its performance with a cross-validation method that excludes each data point once. On the UCF and UIUC datasets, this method produced correct results, however its underlying training procedure and optimization strategy are not capable of handling a vast amount of training data. Many shadow detection methods have been developed using deep neural networks trained with stochastic gradient descent(SGD), which allows for leveraging large amounts of training data.Vicente et al[21] .'s By combining a patch-based CNN with an fully convolutional network that operates on an entire image, the stacked CNN architecture achieved favorable shadow detection outcomes, but it is laborious since the Fully Connected Network (FCN) must first be trained before the patch-CNN can be trained using its predictions. Nguyen et al. [22] introduced scGAN, a method based on Generative Adversarial Networks (GANs), to address the computational expense of using fully convolutional neural networks (FCNs) followed by dense patch-based predictions for shadow detection in test images. Their proposed solution was a parametric conditional GAN architecture, where a generator was trained to produce a shadow mask based on an input RGB patch and a sensitivity parameter. To generate the final shadow mask for an input image, the generator needed to be applied to multiple image patches at different scales, and the resulting outputs were averaged. For the SBU dataset, their method produced good results, although the identification process was computationally expensive during testing. . For shadow detection, our suggested method likewise makes use of adversarial training, but it differs significantly from scGAN in key ways. While our approach utilizes a generator to produce enhanced training images in the RGB space, scGAN applies the generator to generate a binary shadow mask that is conditioned on the input image. In contrast, scGAN employs the discriminator as a regulatory mechanism to encourage global consistency, it is more prominently used in our method to classify shadow pixels. Unlike scGAN, our approach achieves real-time shadow detection without requiring output averaging or post-processing steps. Stacked Conditional GAN [23] is another technique for shadow detection that uses GAN. Nevertheless, this approach needs shadow-free photos to work. A different recent method suggests using contextual information to improve shadow recognition. By leveraging various spatial-directional RNN , our method incorporates both local and global information to enhance shadow detection performance on standard benchmarks. Nonetheless, to further refine the results, a post-processing step utilizing a Conditional Random Field (CRF) is required. In accordance with current trends in data augmentation, we provide a technique to enhance shadow detection using enhanced training samples. Zhang et al. [24] proposed a simple data augmentation strategy that enhances the network's generalization and resilience to adversarial examples by augmenting the dataset with linear combinations of pairs of examples and their corresponding labels. Shrivastava et al. [25] suggested another method that trained a network using adversarial instances. Scientists used adversarial training to create a Refiner network,which takes artificial samples as input and produces more accurate images. The improved examples can be utilized as additional training data to further enhance the network's performance. Similarly, our proposed generator creates realistic images with reduced shadows from the original training photos, which serve as extra training examples for our shadow detector. In contrast to [25], our approach generates new training examples through image manipulation rather than relying on external data sources, where the generated data is a preprocessing step to enrich the training set, The generation of adversarial examples is an integral part of the cooperative training process involving the Discriminator in our approach. Adversarial perturbations are also related to the effects of the shadow Generator changes the input images in order to deceive the shadow detector D-predictions. Net's Adversarialexamples, which a  CGAN uses to do feature engineering, can also be utilized to increase the network's generalization for domain adaption . Kaushal et al. [26] introduced Rapid-YOLO, a one-step shadow detection approach utilizing YOLO V3 that detects shadows through regression, producing bounding boxes for shadow regions. The model incorporates a spatial pyramid pooling(SPP) network to boost accuracy on their proprietary dataset.

## 3  Proposed Methodology

The efficiency of dataset processing has been transformed with the introduction of GANs in generative AI. Large datasets of various and realistic images can be produced using GANs, which cuts down on the time and expense associated with data gathering and annotation. A neural network model called the encoder-decoder architecture is utilised for image processing tasks including picture segmentation and denoising. It consists of an encoder and a decoder that collaborate to recover features from images.  The encoder-decoder architecture has been improved by the introduction of GANs, leading to the creation of more realistic and varied images. Combining these two factors has revolutionised generative AI, enabling the development of more complicated models that can handle challenging computer vision and image processing tasks. To train these models, real-time datasets from various sources have been used.

### 3.1  Encoder-Decoder architecture

Encoder-Decoder architecture commonly used in image processing tasks. The convolutional layers in the encoder phase of the algorithm include increasing numbers of filters, and they are followed by max pooling layers that aid in extracting significant features from the input picture. Convolutional layer with 128 filters makes up the encoder's top layer. The decoder component uses the upsample function to effectively double the feature maps' size by upsampling the encoder's feature maps. Several convolutional layers with a reducing number of filters make up the decoder as well. The 1x1 convolutional layer with a sigmoid activation function, the decoder's last layer, produces a binary mask that represents the detected shadows.

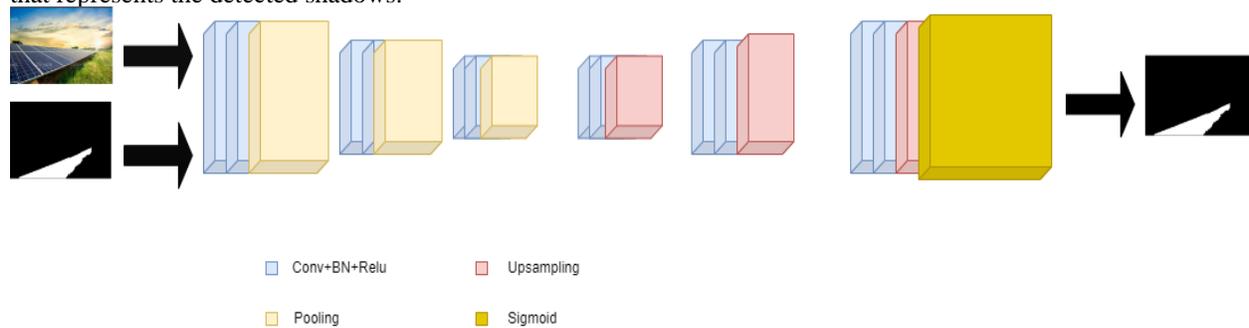

Figure 1: Architecture of Encoder - Decoder

### 3.2  General Adversarial Network architecture

#### 3.2.1  Generator

The generator network trains concurrently with the main network during training. The generator network builds an attenuation mask to reduce the number of shadows in the input image. After applying the mask to the input image, a shadow-free image is created. The output of the generator network is sent to the main network, which has been trained to forecast how well the mask will eradicate shadows. This iterative process continues until the convergence of the generator and main networks yields a satisfactory result. The training procedure incorporates the generator network..

### 3.2.2 Discriminator

The discriminator in the network acts as a binary classifier that assesses whether the input image and the corresponding shadow-free image are genuine. It contains multiple convolutional layers, followed by fully connected layers, and produces a single numerical value that signifies whether the input is probable to be real or fake. To improve the quality of the generated shadow attenuation masks, the generator acquires input from the discriminator, which has been trained to distinguish between authentic and fake inputs. The discriminator's function is crucial in ensuring that the generated masks minimize the presence of shadows while retaining important features of the input image.

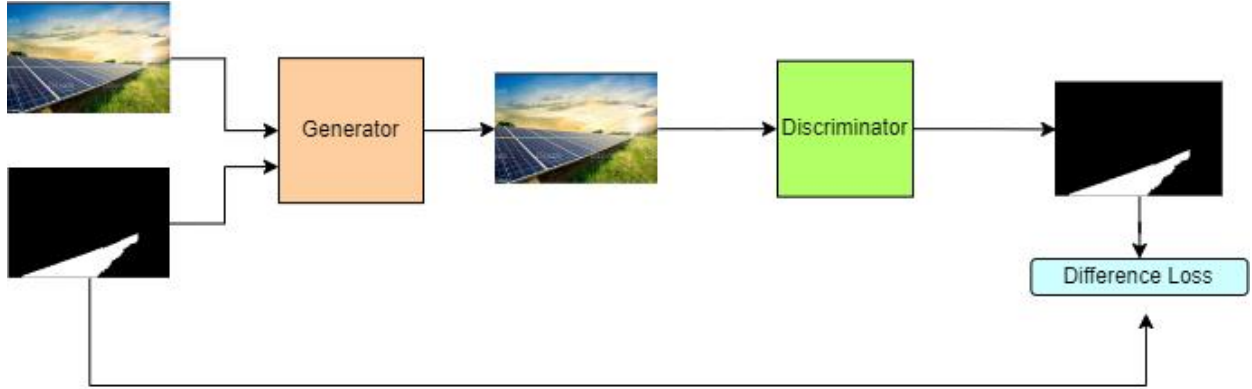

.
Figure 2: Working principle of GAN for shadow detection

### 3.3 Shadow detection using YOLO architecture

Convolutional neural networks are trained to detect the presence of shadows in a picture using the YOLO (You Only Look Once) architecture[27,28]. With excellent accuracy and quick performance, the well-known object detection algorithm YOLO can recognise several items in an image. In the YOLO architecture, a single neural network predicts several bounding boxes and class probabilities for those boxes at once. A dataset of photos with and without shadows is used to train the network to recognise shadows. The model that is produced can then be used to identify shadows in fresh images, which is helpful for tasks like surveillance, driving autonomous cars, and image analysis.

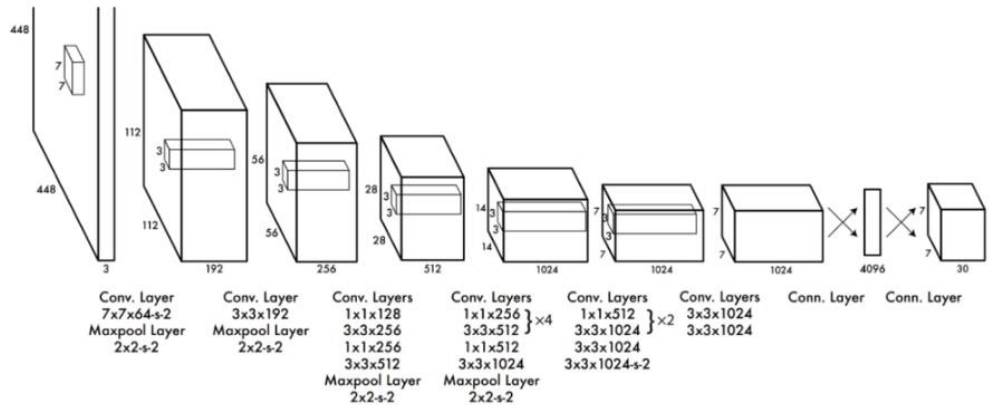

Figure 3: YOLO Architecture

### 3.4 Dataset acquisition

To facilitate data engineering on a larger scale, and introduce variability to the data collection, the data collection process involved a combination of manual real-time data capture and data augmentation techniques, which included applying functions like horizontal flipping, warping, and noise addition to generate these features. For the purpose of facilitating the solution, a total of 250 images were gathered and manually labeled using the Roboflow program. The obtained dataset contained jpg-formatted photos. The photographs will be adjusted to fit each of the cutting-edge models utilised in this work. The dataset was divided into three parts using an 8:1:1 ratio for training, validation, and testing, respectively.. The classes indicated above that were compiled for this task are shown in Table 1 as a preview.

**Table1: Previewing Images Categorized into their Respective Class.**

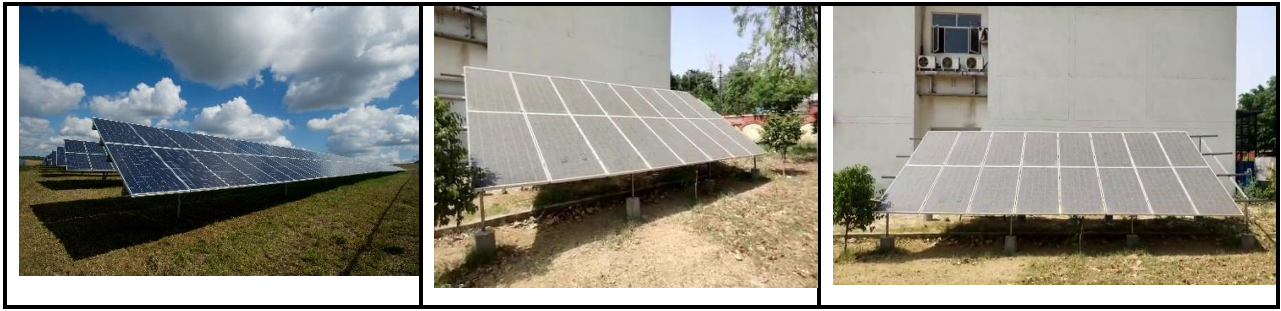

## 4 Results and Discussion

In this section, the Object Detection models are evaluated using various performance metrics curves. Computational cost is measured as the amount of resources utilized by the neural network during training or inference, which helps determine the necessary time and computing power for the NN's operation. Floating-point operations per second (FLOPS) express the number of operations needed to execute a single instance of a specific model, providing insight into the model's computational expenses. The mAP and loss are used to examine the effectiveness of the training process for several object detection models, all of which have undergone 100 epochs of training. The outcomes are presented in Table 1, and the loss and epoch graphs depicted in Table 2 show an inverse correlation, indicating that the implemented models for the given dataset are learning towards convergence. The loss decreases until epoch 80, after which it remains relatively constant. This explains that maximum model learning is achieved between epochs 0 and 80. The Mean Average Precision (mAP) is a widely used performance metric for object detection models. It measures the overall detection accuracy of a model by computing the average precision for each class and then taking the mean overall classes. The map50 and mAP50-95 are variations of the mAP metric that consider the confidence threshold for detection. The mAP50 measures the mAP at a confidence threshold of 0.5, while the mAP50-95 measures the mAP across a range of confidence thresholds from 0.5 to 0.95. Table 1 shows that YOLO v8 has a higher mean average precision (mAP) at a confidence threshold of 0.5 compared to YOLO v5, indicating that it is better at detecting objects with higher confidence. In contrast, YOLO v5 has a lower mAP at a confidence threshold of 0.1, indicating that it may be less effective at detecting objects with lower confidence. The graphs that plot the mAP at different epochs show the performance of the models over the course of the training process. By analyzing these graphs, we can observe how the mAP values change over time and determine the model's overall improvement in accuracy.

In summary, the mAP vs Epoch graphs provides a comprehensive view of the performance of YOLO v5 and YOLO v8, allowing us to compare their ability to detect objects with different confidence thresholds and observe their improvement in accuracy over the training process.

**Table 2. Evaluation of Training Performance for Deep Learning Models with Real-Time Dataset Collection.**

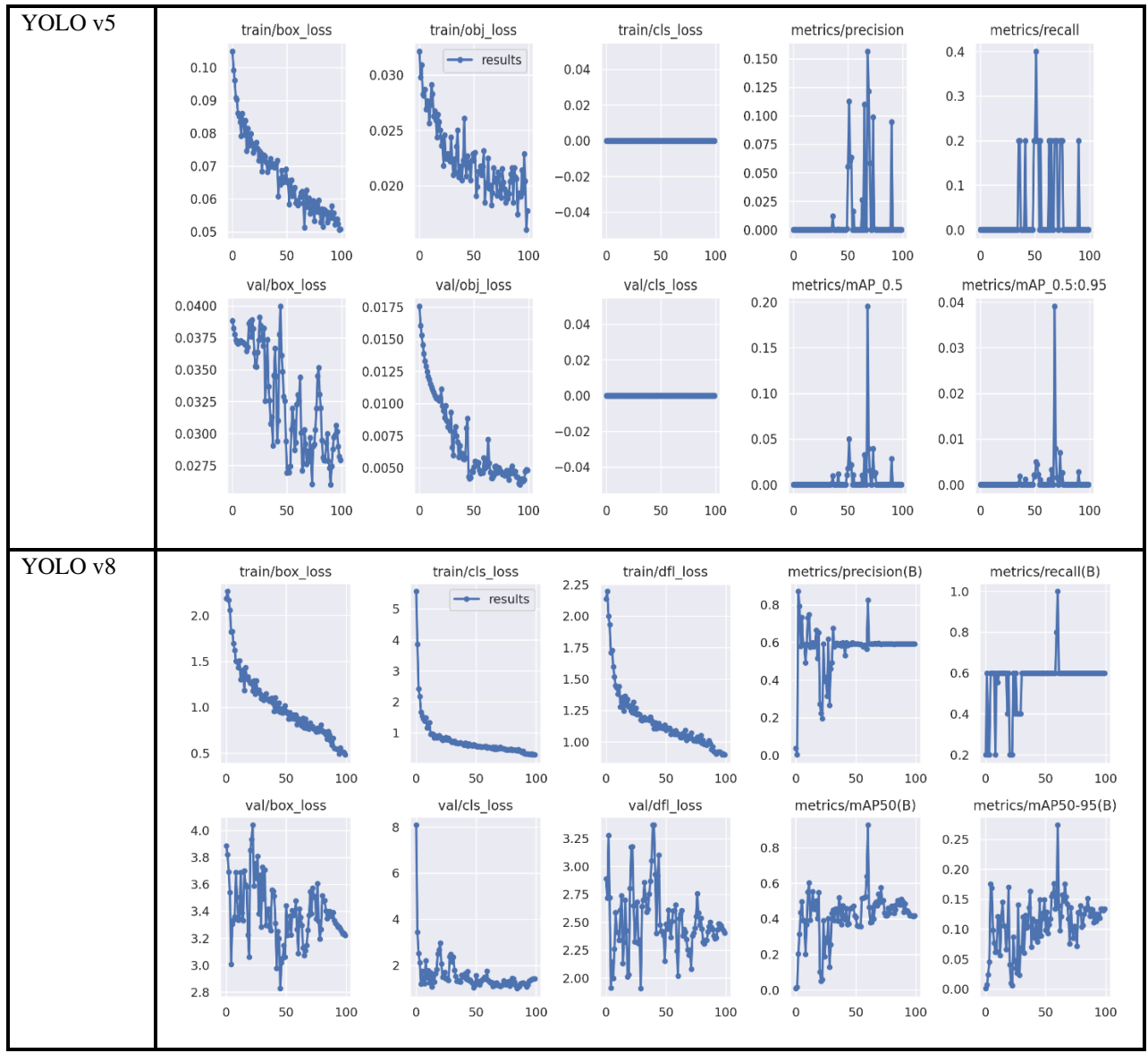

The F1 curve can be used as a valuable tool for assessing the effectiveness of object identification models and can reveal information about the trade-offs between precision and recall at various threshold settings. The FI curve presented in Table 3 examination showed that YOLOv8 outperformed YOLOv5, and this may be attributed to a variety of factors, including the model architecture and training hyperparameters. The precision-recall (PR) curve is a common visualization tool used by object identification models to illustrate the relationship between accuracy and recall at different confidence thresholds. By adjusting the confidence threshold used to accept or reject detections, and then computing precision and recall at each threshold value, the PR curve is created. As it retains good precision at high recall levels, a model with a steeper PR curve performs better. P and R curves can be used as another tool for visualising the effectiveness of object detection algorithms. The P curve displays how precision changes with the number of identified items, whereas the R curve displays how recall varies with the number of detected objects. A model with a high P and R curve will also have a high recall and precision rate.

**Table 3. Graphical Analysis of Performance Metrics during Training of Deep Learning Models with Real-Time Dataset Collection**

| | YOLO V5 | YOLO V8 |
|---|---|---|
| F1 Curve | 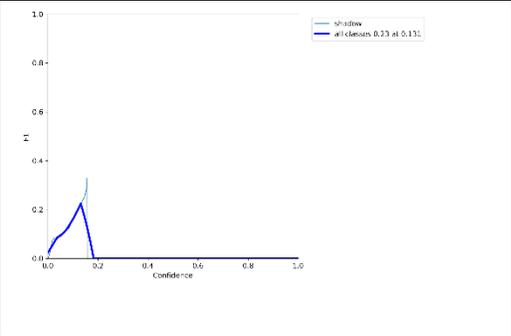 | 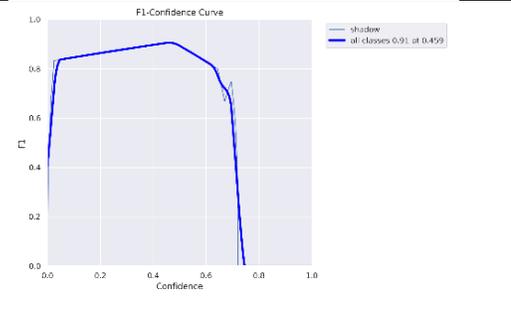 |
| P-R Curve | 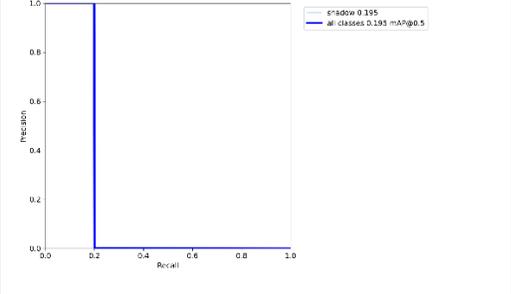 | 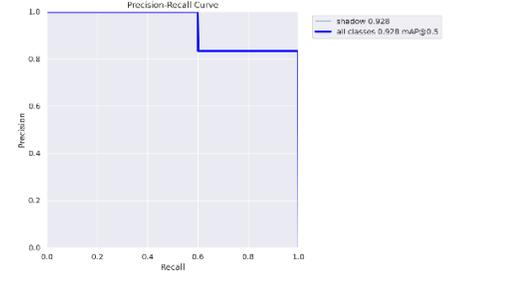 |
| P Curve | 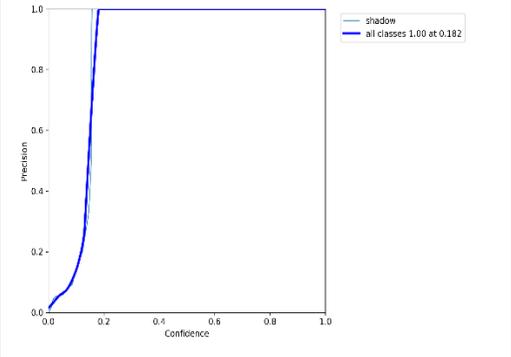 | 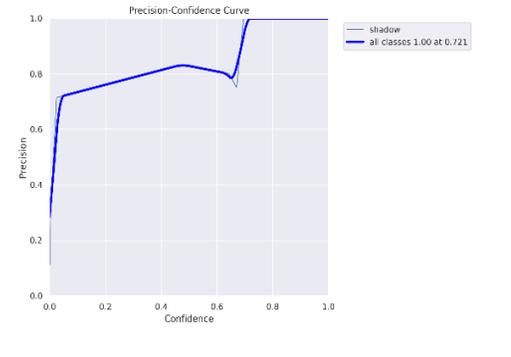 |
| R Curve | 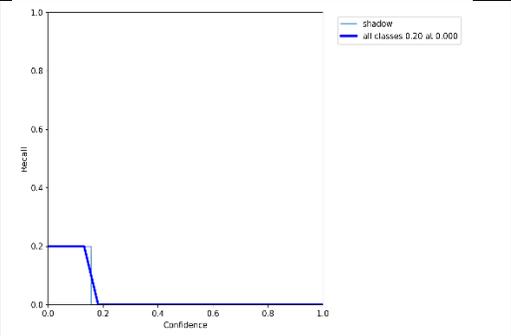 | 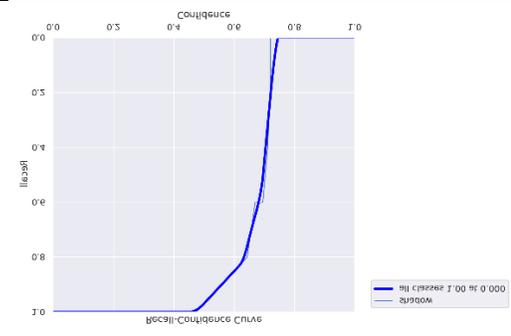 |

Furthermore, from Table 3 when analyzing the precision curve (P curve) and recall curve (R curve), we can see that YOLO v8 outperforms YOLO v5 at different confidence thresholds. YOLO v8 consistently achieves higher precision values than YOLO v5, indicating that it is better at identifying true positives while minimizing false positives. YOLO v8 also consistently achieves higher recall values than YOLO v5, indicating that it is better at identifying all true positives in the dataset.

In summary, the analysis of the P-R curve, P curve, and R curve for YOLO v8 and YOLO v5 demonstrates that YOLO v8 outperforms YOLO v5 in terms of its ability to accurately detect objects across different confidence thresholds. The higher area under the P-R curve, higher R curve, and higher precision and recall values in the P and R curves for YOLO v8 indicate its superior performance compared to YOLO v5.

Table 4. Comparative Evaluation of Performance Metrics across Various Deep Learning Models.

| S. No. | Model | Image Size | Precision | Recall | mAP50 | Flip Flops |
|---|---|---|---|---|---|---|
| 1 | YOLO v5 | 640 X 640 | 0.156 | 0.2 | 0.928 | 16.7G |
| 2 | YOLO v8 | 640 X 640 | 0.826 | 1 | 0.195 | 28.4G |

Based on the information presented in Table 4, the custom YOLOv5s model's performance seems to be relatively subpar on the test set, with low precision and recall, as well as a comparatively low mAP score. In contrast, the YOLOv8s model appears to be performing well on this specific test set, with a high recall and mAP score.

Table 5. Input and Predicted shadow using YOLOv8.

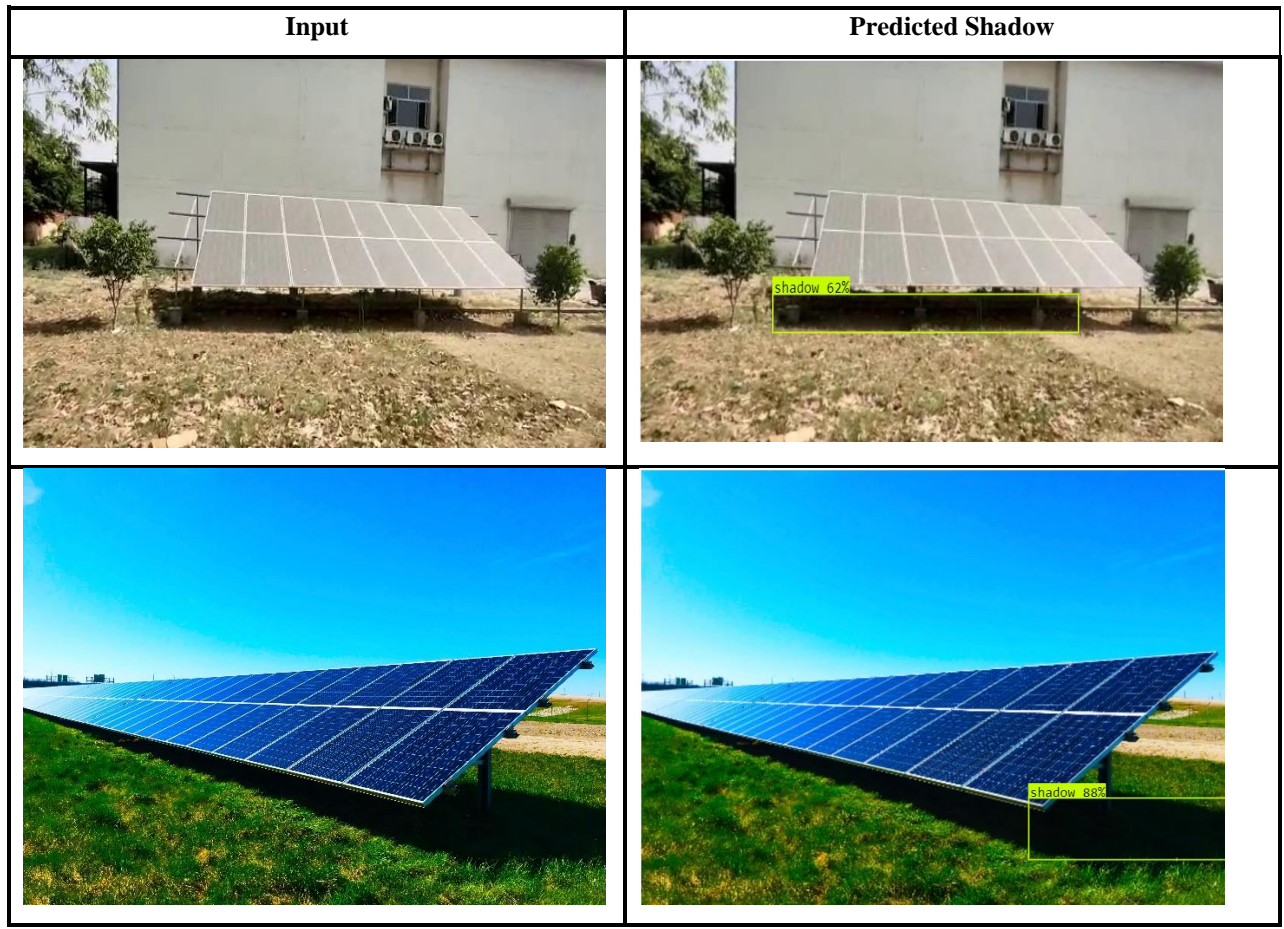

### 4.1 Technical Challenges

These technical errors that were encountered prevented the successful execution of GAN and Encoder Decoder models

- GPU memory error: GPUs are often utilised in deep learning for both inference and training since they are built to handle concurrent calculations. To store the model parameters and intermediate activations during training, deep learning models need a lot of memory. GPU memory errors arise when a model or set of data is too large to fit in the GPU memory. This can occur during model training or inference and result in the training failing or the inference returning a false positive. Using a model with fewer parameters or reducing the batch size can help solve this problem by using less memory.

- Shape mismatch error: Convolutional neural networks (CNNs) are frequently employed in deep learning for image and signal processing jobs where there is a shape mismatch mistake. Convolutional layers typically come first in CNNs, applying filters to the input image to create feature maps, which are then downsampled by pooling layers. For image generating tasks like image segmentation and super-resolution, neural networks frequently employ transpose convolutional layers, also referred to as deconvolutional layers. The filter size, stride, and padding of the convolutional layers have an impact on the shape of the decoder tensor, which must match the shape of the encoder tensor. During training or generation, a shape mismatch error may occur if the encoder tensor's shape is incompatible with the convolutional and transpose processes. This can be resolved by adjusting the filter size, stride, padding, or the number of layers in the network to ensure the shapes are compatible.

## 5 Conclusion

In conclusion, computer vision techniques have become increasingly popular in shadow detection. This study evaluated the performance of GAN, encoder-decoder neural networks, and object detection methods such as YOLOv5 and YOLOv8 for detecting shadows. While object detection methods proved to be efficient for shadow detection, they may not provide the same level of accuracy as GAN and encoder-decoder neural networks. The study found that YOLOv8 outperformed YOLOv5 in all metrics, making it a preferred choice for shadow detection. However, technical errors prevented the implementation of GAN and encoder-decoder neural networks, highlighting the need for further research in this area to address these challenges and explore the potential of these more precise techniques for shadow detection. Overall, this study shows that object detection methods like YOLOv5 and YOLOv8 can be useful for detecting shadows, although the results may not be as accurate as those obtained through the use of GAN and encoder-decoder neural networks. Additionally, YOLOv8 was found to be a superior choice for this task. Further research is necessary to investigate the potential of GAN and encoder-decoder neural networks in shadow detection and to address the practical challenges associated with their implementation. This study provides important insights into the use of computer vision techniques for shadow detection and highlights areas for future research.